\documentclass[11pt,a4paper]{article}

\usepackage[margin=1in]{geometry}
\usepackage[utf8]{inputenc}
\usepackage[T1]{fontenc}
\usepackage{booktabs}
\usepackage{graphicx}
\usepackage{xcolor}
\usepackage[normalem]{ulem}
\usepackage[most]{tcolorbox}
\usepackage{enumitem}
\usepackage{fancyvrb}
\usepackage{fvextra}
\usepackage{amsmath,amssymb}
\usepackage{float}
\usepackage[breaklinks,pdfauthor={Marios Adamidis, Danae Katrisioti, Yannis Tzitzikas, Emmanuel Stratakis},pdftitle={It's not the Language Model, it's the Tool: Deterministic Mediation for Scientific Workflows},pdfsubject={Typed mediation for reproducible scientific analysis},pdfkeywords={language models, reproducibility, tool mediation, MCP, scientific workflows}]{hyperref}
\usepackage{url}
\usepackage{authblk}

\tcbuselibrary{breakable}

\newcommand{\tool}{FORTHought}

\begin{document}
\title{It's not the Language Model, it's the Tool: \\
Deterministic Mediation for Scientific Workflows}
\author[1,2]{Marios Adamidis}
\author[1,2]{Danae Katrisioti}
\author[3,4]{Yannis Tzitzikas}
\author[2,5]{Emmanuel Stratakis}

\affil[1]{Department of Materials Science and Technology, University of Crete, Heraklion, Greece}
\affil[2]{Institute of Electronic Structure and Laser, FORTH, Heraklion, Greece}
\affil[3]{Computer Science Department, University of Crete, Heraklion, Greece}
\affil[4]{Institute of Computer Science, FORTH, Heraklion, Greece}
\affil[5]{Department of Physics, University of Crete, Heraklion, Greece}

\date{May 13, 2026}
\maketitle

\begin{abstract}
Language models can produce convincing scientific analyses, but repeated generations on the same data do not guarantee the same result. A researcher may regenerate an identical query and receive a different fit, a different peak position or a different analysis procedure, without an obvious way to decide which output to trust. We propose {\em typed mediation}, a pattern in which the model orchestrates deterministic tools rather than generating analytical code. Each tool encodes one researcher's exact procedure for one instrument, ported through structured interviews. The model selects which tool to call and with what parameters. The tool produces the result. Regeneration does not change it. We evaluate this claim by running the same photoluminescence analysis on four platforms, including three commercial foundation models, four times each with the same prompt. The typed tool produces identical results across all runs. The commercial platforms either vary in numerical output and analytical methodology across runs, or fail to produce valid results on the task. We deploy this pattern on two instruments serving users over approximately six months, with very positive user feedback.
Both cases are very challenging: they
involve proprietary binary formats and per-seat licensed software, which force the tool to remain on local infrastructure alongside the data and the instrument it operates. We argue that deployment topology is not just a preference, but a structural requirement of scientific tool mediation.
The result is a practical pattern for deploying language models
in scientific workflows where reproducibility is mandatory,
reducing analysis time from weeks to minutes
while guaranteeing identical outputs across runs.
\end{abstract}

\section{Introduction}

\noindent
\textbf{Motivation: Reproducibility First.}
The integration of language models into experimental data analysis is accelerating, and with it the assumption that the model's output is reliable enough to act on.
For instance, 
consider the case of a researcher,
who after a long day of measurements, 
returns to the laptop to analyze a messy photoluminescence spectrum file\footnote{A photoluminescence spectrum records the light emitted by a material after optical excitation, as a function of wavelength. It is a standard characterization technique in semiconductor physics.}.
The researcher uploads the file to the AI assistant they have used many times before and found efficient at producing what they consider passable. The model proceeds to work on the file via code execution and rapidly arrives at a result that appears well crafted and articulated. From the peak fit to the plot and the analysis, the output validates the researcher's hard work. There is only one simple issue: the researcher forgot to paste the right labels for the measurements, so they edit the query and press the regenerate button. This time, the peak position has shifted by 2~eV, the fit is just a bit different. Neither answer is obviously wrong. What is the principled way for the researcher to act at 9pm on a Thursday?

One important observation
is that
this failure 
\emph{does not go away with a more powerful model.} The model's view of the project's semantics is governed by a generic system prompt, not tailored to any specific researcher's workflow. The model is capable enough to consistently produce convincing results. What it cannot do is produce the same result twice. The gap is determinism, and more model capability does not provide it~\cite{he2025determinism}.
Reproducibility is not a luxury in science; it is the standard that separates real discovery from mere claim.

This widespread issue appears as well in the low adoption of agents in academic research and medicine, sitting at low single digits of real-life deployment as shown in Anthropic's recent report~\cite{anthropic2026agents}. These fields have been investigated greatly for the potential of AI-related integrations, but a lack of technical know-how, combined with the marketing pressure that the model is already capable enough on its own, has made it impractical for unsupported researchers to graduate from zero-shot prompting to task-strengthened, specialized harnesses~\cite{kitchin2025programming}. 

\smallskip
{\bf Motivation: The Need for Privacy.}
A privacy concern also arises from the interaction. Many types of laboratory data appear uniformly consistent to the human eye, dampening the researcher's instinct towards what they will share with the cloud provider. A file containing a bad measurement is virtually indistinguishable from one that will lead to an impactful publication. Sharing raw data and results with the cloud provider creates a non-zero probability that the data shapes the next checkpoint's performance in that specific niche~\cite{balloccu2024leak}. Cutting-edge scientific output has higher probability of being valuable on later rounds of training when compared to trivial information extracted from the web.

\begin{figure}[H]
    \centering
    \includegraphics[width=0.8\linewidth]
    {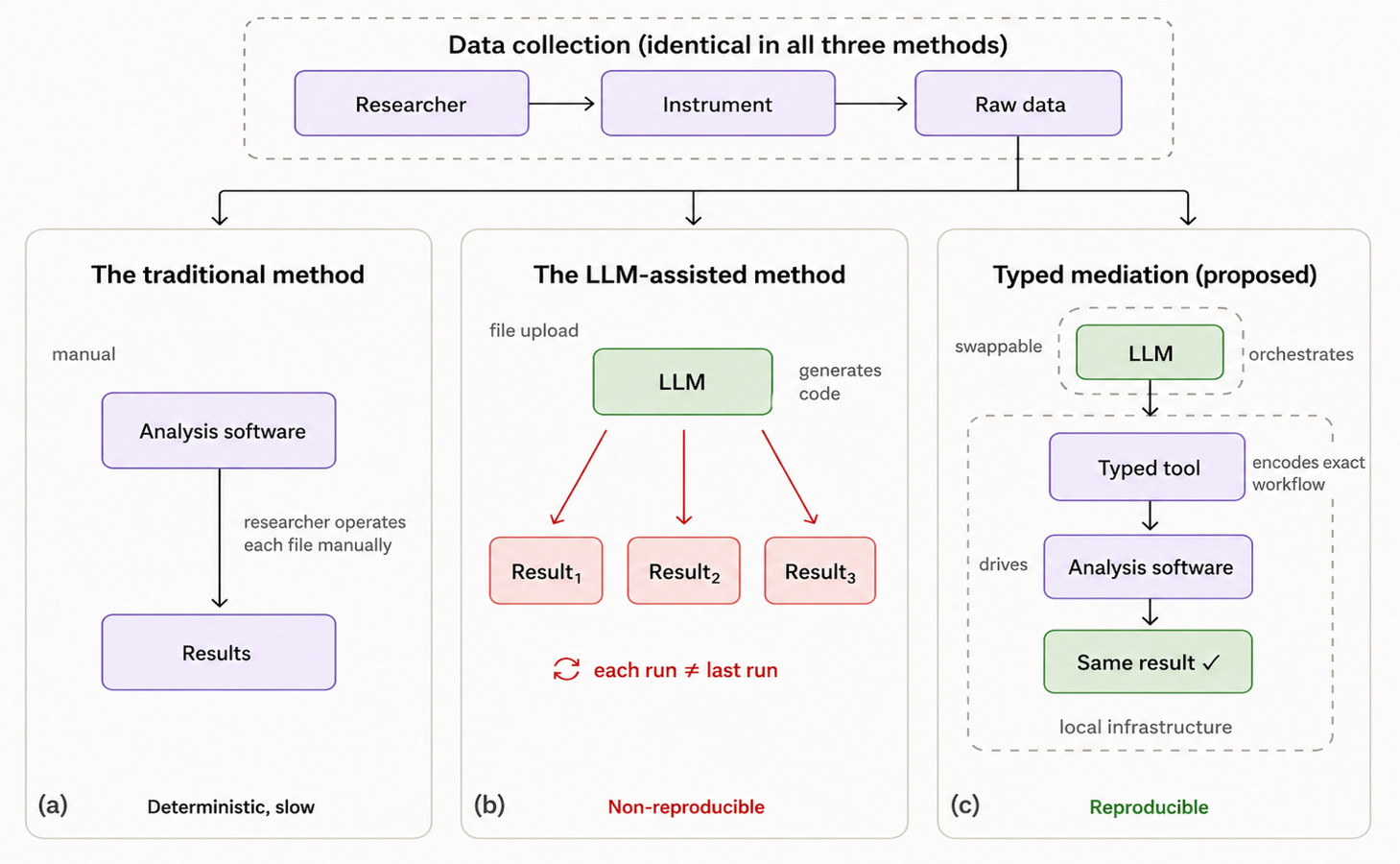}
    \caption{Three approaches to scientific data analysis. Data collection is identical in all cases. (a)~Traditional: the researcher manually operates analysis software. (b)~LLM-assisted: the model generates analytical code, producing different results on each run. (c)~Typed mediation (proposed): the model orchestrates a deterministic typed tool that encodes the researcher's exact workflow on local infrastructure.}
    \label{fig:Overview}
\end{figure}

\begin{figure}[H]
  \centering
  \includegraphics[width=0.99\textwidth,keepaspectratio]
  {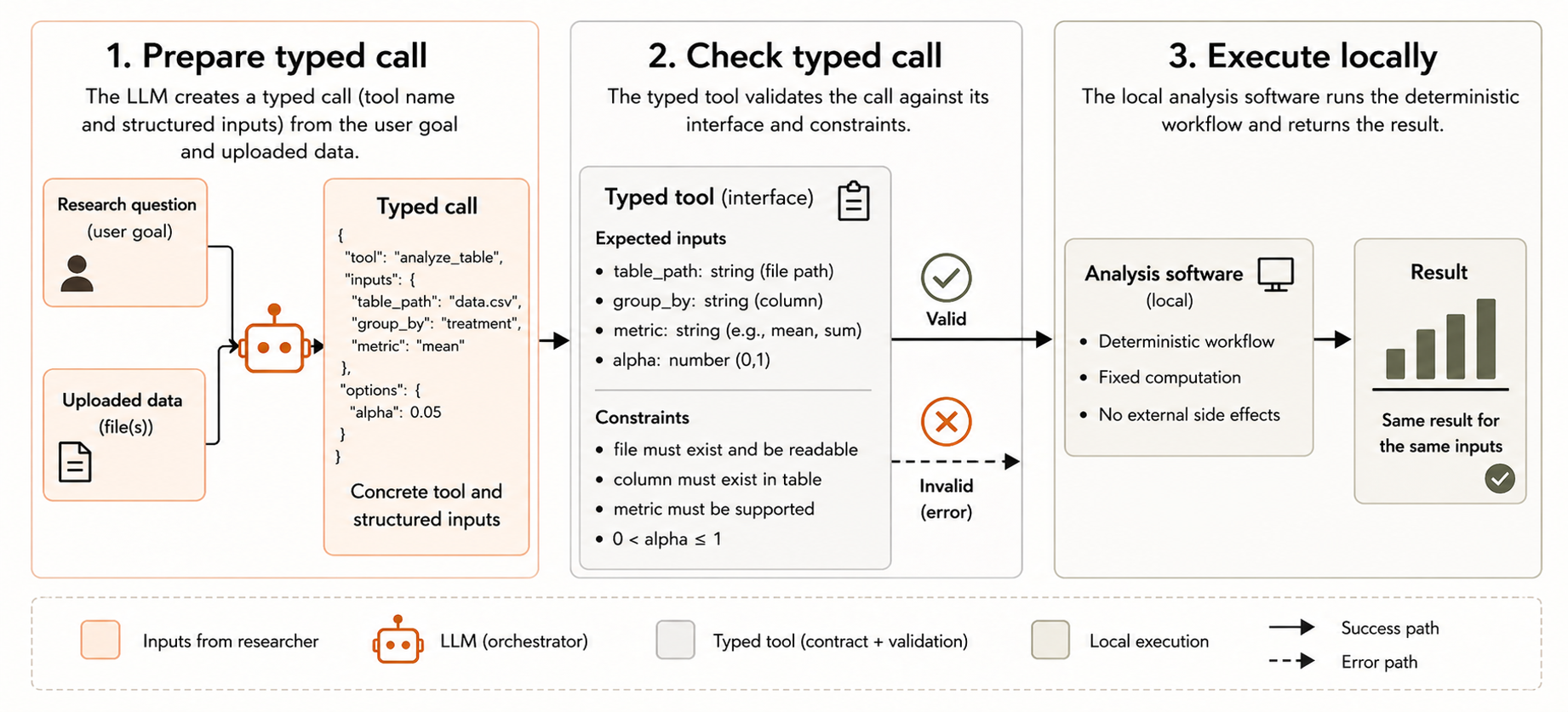}
  \caption{Typed mediation architecture. The model orchestrates tool calls through a typed schema. The tool encodes the scientist's workflow and drives licensed software on local infrastructure.}
  \label{fig:architecture}
\end{figure}

\noindent
\textbf{Approach.}
For tackling the above 
requirements,
i.e. reproducibility, and privacy,
we propose an alternative way
to leverage models.
In particular, we propose encoding the exact manual task of the researcher behind a tool surface. 
Then the model orchestrates. 
Subsequently, the tool produces deterministic results that reproduce across regenerations.
In this way, verification becomes quick, and the model's reasoning capacity goes to the parts of the work that actually need it.
Figure~\ref{fig:Overview} gives an overview of the proposed approach.
In the left part we see
the traditional method, 
relying on particular software
and manual operation.
In the middle part
we see the current trend,
i.e. to use an LLM 
for the analysis, 
but 
since LLMs operate
stochastically, 
their output varies,
therefore 
they derive  
non-reproducible
results.
The right part
illustrates our proposal,
i.e. to leverage
{\em both}
the existing software
and the LLM
for getting 
reproducible results.
We could say,
that we "wrap" existing
software
in a way
that is convenient
for the language model
to orchestrate
deterministic workflows.
This raises the question we address in this paper:
When the workflow lives in the tool rather than the model, how much does model choice matter?
A more refined illustration is given in  Figure \ref{fig:architecture}.

\smallskip
\noindent
\textbf{Evaluation.}
We have realized this  approach
as a platform, called \tool,
we have deployed it 
in a research center,
and the platform 
is in use
by eleven researchers,
across multiple instrument workflows,
for a period of 6 months.

The users  have provided
very positive feedback,
and  \tool\  has now become part of their daily routine.
In this paper, we shall
focus on 
(a) the two deployments with the largest documented impact: 
a photoluminescence analysis pipeline and a scanning electron microscopy workflow,
and 
(b) on the evaluation of  {\em reproducibility} by running the same photoluminescence analysis on our platform and three commercial foundation models, four times each with the same prompt and the same data.

\smallskip
\noindent
\textbf{Contributions.}
In summary, our work makes four main contributions:
\begin{enumerate}
    \item we describe a {\em typed-mediation pattern} in which both the human and the language model address the laboratory software via the same typed interface, placing the deterministic core of the workflow in the tool rather than the model,
    \item we present {\em two real applications} ported through structured interview sessions with the researchers who own the workflows, deployed and actively used by researchers who verify outputs quickly and feed corrections back into the tools,
    \item we evaluate reproducibility by running the same analysis on four platforms four times each and show that the typed tool produces identical results across runs while code-generating approaches vary in both output and analytical methodology, and
    \item we argue that deployment topology is a structural requirement of scientific tool mediation, 
driven by both privacy concerns and the licensing constraints of most laboratory software, which together force the tool to live alongside the data and the instrument it operates.
\end{enumerate}

The rest of this paper is organized as follows.
Section~\ref{sec:relatedWork} discusses related work.
Section~\ref{sec:Our} describes our approach, i.e. the typed mediation pattern.
Section~\ref{sec:deploymentCases} presents two deployment cases.
Section~\ref{sec:evaluation} evaluates reproducibility across platforms.
Section~\ref{sec:CR} concludes and outlines future directions.

\section{Related Work and Novelty}
\label{sec:relatedWork}

Deployed laboratory agents remain rare. Hellert et al. report a production deployment at a synchrotron user facility, where an agentic framework manages real-time operations across more than 230,000 control channels~\cite{hellert2025accelerator}. Vriza et al. evaluated code-generating agents at a national X-ray nanoprobe facility and found that performance on the same task varies sharply across models~\cite{vriza2026instruments}. Xie et al. demonstrated LLM-driven control of a scanning photocurrent microscope, illustrating both the potential and the fragility of code-generating approaches for instrument automation~\cite{xie2026labcontrol}. Ciss\'{e} et al. tested five reasoning models as scientific optimizers over twenty repeat runs each and observed that a single model can produce outlier results below half its own average performance~\cite{cisse2026bora}. Cui and Alexander ran 480 attempts across six models on the same data analysis task and found considerable variation in analytical results even under identical configurations~\cite{cui2026sameprompt}. These results confirm that within-model variance is a structural property of code-generating approaches, not a failure of any particular model.

Tool-mediated architectures address this by restricting the model to validated tool calls. Yang et al. report 100\% tool selection consistency across three models at maximum sampling temperature~\cite{yang2026chatspatial}. Xu et al. achieve 100\% specification-level reproducibility when the model is limited to routing decisions~\cite{xu2026reproducibility}. Pan et al. demonstrate MCP-based tool mediation entering scientific cyberinfrastructure at the national laboratory scale~\cite{pan2025mcpscience}. Strickland et al. formalize this principle as schema-gated orchestration, where nothing executes unless it validates against a machine-checkable specification~\cite{strickland2026talkfreely}. Doshi et al. propose extending MCP with structured labels for capabilities, confidentiality and trust, enabling deterministic enforcement of safety constraints at tool boundaries~\cite{doshi2026verifiablysafe}. Deng et al. present a skill-centric framework for autonomous operation across ten categories of precision instruments, where reusable operational and analytical skills connect physical sample handling with scientific interpretation~\cite{deng2026owlauraid}.

Our contribution is operational rather than architectural. We take the schema-gated principle and apply it to a specific problem that none of these systems address: encoding one researcher's exact manual workflow, extracted through structured interviews, as a typed tool that any model can orchestrate.

\section{The \tool\ Approach}
\label{sec:Our}

At first we should clarify terminology.
The key concepts and their terms and descriptions, 
are given in Table~\ref{tab:terminology}.
The reader is suggested to read it,
to avoid ambiguities.

\begin{table}[H]
\caption{Key terminology of \tool}
\label{tab:terminology}
\centering
\footnotesize
\begin{tabular}{p{0.18\textwidth} p{0.75\textwidth}}
\toprule
\textbf{Concept} & \textbf{Definition} \\
\midrule
\emph{Software} & Any application a human operates directly: a spreadsheet program, a licensed instrument suite, a standalone script. Designed for human interaction. \\
\addlinespace
\emph{Tool} & A function exposed to the language model through a programmatic interface. The model calls it; the user does not operate it directly. \\
\addlinespace
\emph{Schema} & A machine-readable specification of a tool's expected inputs and outputs: parameter names, types, allowed values and return format. It acts as the contract between the model and the tool. \\
\addlinespace
\emph{Typed tool} & A tool whose interface is defined by a strict schema. A call that violates the schema is rejected before execution. A call that conforms executes the same deterministic procedure every time, regardless of which model issued it. \\
\addlinespace
\emph{MCP} & The Model Context Protocol~\cite{anthropic2024mcp}: an open standard that exposes typed tools to language models. It defines how a model discovers available tools, reads their schemas, and issues calls. \\
\addlinespace
\emph{Skill} & A configuration document that defines the model's behavior for a specific instrument or researcher workflow. It specifies which tools are available, what the researcher's preferences are, and how results should be presented. Each deployment case has its own skill.
An example in given in Appendix \ref{sec:Skills}.
\\
\addlinespace
\emph{Workflow specification} & The output of the structured interview process: the complete set of analytical decisions a researcher applies to their data fitting model, spectral window, quality thresholds, preprocessing steps. Each typed tool encodes exactly one workflow specification. 
An example in given in Appendix \ref{sec:Interview}.
\\
\bottomrule
\end{tabular}
\end{table}

We could summarize the entire approach as follows:
\noindent
\begin{tcolorbox}
Each {\em typed tool} encodes a {\em workflow specification} extracted through {\em structured interviews} with the researcher, 
{\em wraps} the operations of {\em existing licensed software}, and is {\em exposed to the model} via MCP.
A per-instrument {\em skill} tells the model 
which tools are available and how the researcher expects results to be handled. 
The model calls the tool, and the tool drives the software.
\end{tcolorbox}

\subsection{Typed Mediation}
\label{sec:TypedMediation}

\noindent
{\bf Typed Mediation.}
We use the term {\em typed mediation} to refer to a pattern in which the language model does not perform the analysis itself, but selects and invokes a deterministic tool that encodes the researcher's exact procedure through a typed schema.

\smallskip
\noindent
{\bf Determinism.}
Typed mediation places a deterministic tool between the software that the researcher was manually using and the model. A system prompt and an accompanying per-experiment {\tt Skill.MD} file define the model's appropriate behavior towards the task relevant to the user's profiled requests. 
An example of a "skills file",
is given in Appendix \ref{sec:Skills}.
The model will decide which tool to execute and with what parameters as input. The tool then proceeds to deliver the precise results back to the model, which is instructed to just serve or reason upon the results. This interface follows the Model Context Protocol~\cite{anthropic2024mcp}, which is an open standard that exposes each tool through a typed schema specifying its inputs, outputs and execution requirements. The tool does not act as a generic API wrapper. It encodes one researcher's exact procedure for one instrument or experiment, for example what despiking parameters they use, the spectral window they fit over and the peak boundaries they integrate between. Simply, the model picks which tool to call, and the tool itself already knows how the work is done.

\smallskip
\noindent
{\bf Architecture Overview.}
Figure~\ref{fig:architecture} summarizes the typed mediation architecture. The model operates above the typed interface and is replaceable. The tool sits below it, anchored to the same machine as the licensed software and the instrument.

\smallskip
\noindent
{\bf Why we need typed mediation.}
By design, language models are stochastic, which is both a curse and a blessing. It allows them to appear creative and energetic in domains where verbatim recall would fail. But in applications such as scientific analysis, the methodology used directly affects the immediate result and the relationship between data produced in different sessions. A deterministic tool surface allows for the creativity of the model to be utilized towards detecting vital insights from within the process and provide better analyses using the tool output as grounding. The model's reasoning traces get occupied by science-related reasoning rather than what code to execute to please the user's requirements. Recent work confirms this empirically: 100\% tool selection consistency across three models set at maximum sampling temperature~\cite{yang2026chatspatial}, and 100\% specification-level reproducibility when the model is restricted to routing decisions~\cite{xu2026reproducibility}.

The other path is to let the model generate the code required to satisfy the user's demands, call generic APIs or reason about the workflow itself. This is the current approach of most users, as reported by Anthropic~\cite{anthropic2026agents}. In a recent study, Vriza et al. evaluated this directly at a national user facility (X-ray nanoprobe + robotic thin-film). On the hardest task, performance ranged from 0\% on weaker models such as GPT-4o-mini, to 25\% on Claude 3.5 Sonnet with high variance, to 100\% on OpenAI's O3 model~\cite{vriza2026instruments}. This apparent spread on the same task casts doubt on even the most seemingly consistently performant models, because it proves that the process itself is volatile.

That volatility disappears when the model is restricted to selecting which tool to call. The typed schema accepts valid parameters and executes the same procedure regardless of which model issues the call. But the tool itself cannot move. Proprietary binary formats open only inside their licensed application, per-seat licenses bind that application to one workstation, and instrument-specific calibration files encode settings that apply to one physical device~\cite{diao2026integrating}. A typed interface exposed through MCP resolves this asymmetry: the tool remains where the work lives, and any model that can issue a valid call can orchestrate it, from bench instruments to national-scale cyberinfrastructure~\cite{pan2025mcpscience}.

\subsection{Problem Statement and Approach Formally}
\label{sec:Formal}

\noindent
{\bf The problem of stochastic analytical variance.}
Let $D$ be a scientific dataset and $P$ be a natural language prompt describing an analytical procedure. 
In a code-generating LLM workflow (the "LLM-assisted method"), the analysis is treated as a stochastic function $f_\theta$ parameterized by the model weights $\theta$.
The output $R$ is generated as:
$R_{i} = f_{\theta}(D, P, \tau, s_i)$
where $\tau$ is the sampling temperature and $s_i$ is the random seed for run $i$. 
The problem is defined by the {\em variance of methodology} 
$\text{Var}(f_{\theta}(D, P)) \neq 0$,
i.e. even when $D$ and $P$ remain constant, $R_i \neq R_j$ because the model reconstructs the analytical methodology 
(e.g., peak fitting models, background subtraction, spectral windows) 
stochastically at runtime. In a scientific context, this lack of determinism prevents $R$ from being a "verifiable scientific result".

\ \\
{\bf Proposed Solution.} 
We propose a transformation where the LLM is restricted from generating code and is instead limited to {\em parameterizing} a deterministic tool.
Let $T$ be a {\em typed tool} that encodes a fixed analytical workflow $W$.
The tool is governed by a schema $S$, which defines the valid input space $\mathcal{X}$.

In particular, the workflow $W$ is defined by the {\em interview} 
(that extracts the researcher's methodology)
and implemented 
(with the assistance of a developer)
in the tool code ($T$),
while the {\em skill file} defines $S$ (the schema),
i.e.  it tells the model how to call the tool, 
not what the tool does internally
(we could say that the skill file is the menu, not the kitchen).
To summarize, the process is:
{\em Interview} $ \xrightarrow{\text{what analysis to do}} W \xrightarrow{\text{code impl.}} T$,
and
{\em Skill file} $ \xrightarrow{\text{llm}}$ {\em interface} $S$ of $T$.

\noindent
According to this approach, the process is decomposed into two distinct phases: 
\begin{itemize}
	\item  Orchestration (Stochastic): The LLM acts as a {\em mediator} $m_\theta$ that maps $(D, P)$ to a set of parameters $x \in \mathcal{X}$ 
	  such that $x$ satisfies
	 the constraints of $S$, i.e. we can write $x = m_{\theta}(D, P, S)$.
	\item  Execution (Deterministic):  
	The tool $T$ executes the workflow $W$ using parameters $x$,
	  i.e. $R = T(D, x)$.
\end{itemize}
\noindent
This  {\bf guarantees reproducibility} since $T$ is a non-stochastic function implemented in local infrastructure, 
and thus it satisfies the property: 
$\forall i, j : x_i = x_j \Rightarrow T(D, x_i) = T(D, x_j)$.
By enforcing $x$ through a  {\em typed schema} (Schema-Gated Orchestration), 
the system ensures that as long as the model's tool-selection $x$ is consistent, 
the scientific result $R$ is perfectly reproducible. 
This shifts the burden of "correctness" from the model's creative generation to a validated, researcher-defined tool surface.

\noindent
{\bf Summary of Formal Contributions} \\
$\bullet$ Separation of Concerns: Decoupling the reasoning (LLM) from the computation (Typed Tool) \\
$\bullet$ Topology Constraint: Formalizing that $T$ must reside on local infrastructure due to proprietary licenses and data privacy, while $m_\theta$ may reside in the cloud. \\
$\bullet$ Zero-Variance Methodology: Ensuring $\sigma^2 = 0$ for the analytical procedure across multiple model regenerations. 
The deeper finding is that a {\em tight enough schema} makes the LLM's parameter selection deterministic too. 
For example if $b$ is the intensity exponent that we are looking for, 
with our approach we get $\sigma^2_b \approx 0$.  We could further analyze,
and define: $\sigma^2_{total} = \sigma^2_{mediation} + \sigma^2_{execution}$, 
where $\sigma^2_{execution}=0$ by construction and $\sigma^2_{mediation} \approx 0$,  empirically when $S$ is sufficiently constrained.

\section{Deployment Cases of \tool}
\label{sec:deploymentCases}

As stated earlier,
here we describe two deployment cases,
that have significant impact.
A few statistics
about these two cases,
i.e. number of typed functions,
users,
logged sessions,
time (i.e. before and after \tool),
validation,
and other constraints,
are given in 
Table~\ref{tab:deployment}.

\begin{figure}[H]
  \centering
    \includegraphics[width=\textwidth,keepaspectratio]{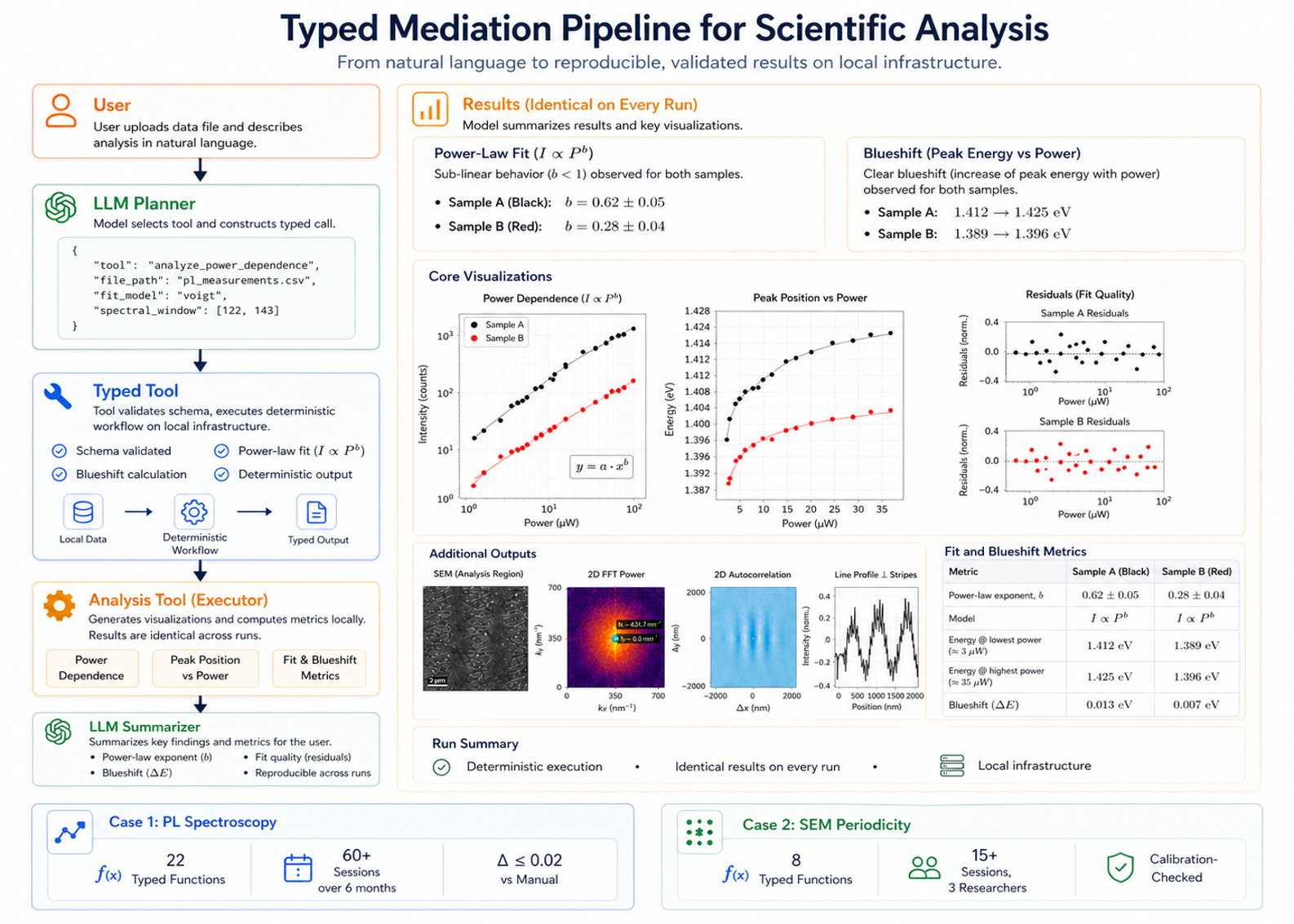}
  \caption{Typed mediation pipeline applied to photoluminescence spectroscopy. Left: the user describes the analysis in natural language; the model selects the tool and constructs a typed call; the tool validates the schema, executes the deterministic workflow on local infrastructure and returns identical results on every run. Right: core visualizations and fit metrics produced by the tool. The bottom strip summarizes both deployment cases (PL spectroscopy and SEM periodicity analysis).}
  \label{fig:pipeline}
\end{figure}

\begin{table}[H]
\caption{Deployment summary. Both tools have been in continuous use for approximately six months.}
\label{tab:deployment}
\centering
\small
\begin{tabular}{lcc}
\toprule
 & \textbf{PL spectroscopy} & \textbf{SEM periodicity} \\
\midrule
Typed functions        & 22         & 8 \\
Active users           & 1 primary + group & 3 \\
Logged sessions        & 60+        & 15+ \\
\midrule
Manual time            & $\sim$2 weeks & $\sim$2 days \\
Tool time              & minutes    & minutes \\
\midrule
Validation             & $\Delta \leq 0.02$ & calibration-checked \\
Local constraint       & binary format, & device-specific \\
                       & per-seat license & calibration \\
\bottomrule
\end{tabular}
\end{table}

\subsection{Photoluminescence Analysis}
\label{sec:Photo}

The first deployment case is a photoluminescence analysis pipeline for semiconductor thin films. The proprietary analysis software stores each excitation power level as a separate workbook inside a binary project file. A typical measurement campaign produces more than twenty such workbooks, each sharing the same wavelength axis but containing a different excitation power, meaning a different emission spectrum. The researcher must open each one, apply the same sequence of operations, extract the results and combine them into a single notebook. The initial tool automated this merging step. Figure~\ref{fig:pipeline} illustrates the full pipeline: the researcher describes the analysis in natural language, the model constructs a typed tool call, and the tool executes the workflow deterministically on local infrastructure.

The tool became useful, but it was not yet correct. To move from automation to fidelity, we conducted a structured interview of thirty questions across two sessions with the researcher who owns the workflow. 
An example in given in Appendix \ref{sec:Interview}.
Each answer refined the tool. The fitting model was Voigt (a convolution of Gaussian and Lorentzian components that better captures peak shape in noisy spectra), not the pure Lorentzian we had assumed. The spectral window adapts per power level, narrowing around the emission peak as it shifts with excitation power, rather than staying fixed. The quality threshold is approximately $R^2 \geq 0.90$, evaluated visually, not the 0.95 cutoff we had implemented. Peak intensity is read as raw counts at the maximum, not as integrated area. For an example case, the power dependence uses two separate fits split at a saturation boundary near $10~\mu\mathrm{W}$, not a single allometric fit. None of these details were documented. They existed in the researcher's routine, refined over four years of doctoral work. The final tool encodes this routine across twenty-two typed functions. The right panel of Figure~\ref{fig:pipeline} shows the core visualizations and fit metrics that the tool produces from a single run.

The raw measurement files use a proprietary binary format. They open only inside the licensed desktop application. The license is bound to one workstation. Institutional policy also prevents uploading raw experimental files to external services. This is the normal condition of many experimental laboratories. The model cannot open the file, cannot run the licensed software and cannot ask the researcher to export data she is unwilling to share. The tool has to run on the same workstation as the data and the software.

We did not build tools that imitate a result. We ported the exact workflow the scientist was already using into a typed tool that the model orchestrates.

Manual analysis of one measurement campaign previously took approximately two weeks. The time was spent on repeated file handling across twenty-one power levels. Each level required the same sequence of operations inside the proprietary application. The ported workflow executes in minutes. The exponent from the automated split power law fit matches the researcher's manual result within $\Delta \leq 0.02$. Regeneration on the same data returns the same numbers every time. The researcher can now verify the output in minutes because she knows what the correct answer should look like.

The researcher is a fourth-year doctoral student working on optical spectroscopy of two-dimensional materials. She was initially skeptical because her existing workflow already worked. After the tool reproduced her method, she began using it regularly. Over approximately six months, she has accumulated more than sixty interaction sessions, including more than eleven sessions on the spectral analysis model. She has also introduced the platform to members of her research group. The deployment is now part of her working routine.

\subsection{Scanning Electron Microscopy}
\label{sec:Scanning}

The second case is a scanning electron microscopy workflow for periodic structure analysis on laser-processed surfaces. The tool encodes the calibration of one specific microscope, including pixel scale, magnification mapping and detector configuration. A different instrument of the same model would require different values.

The porting followed the same method. The researcher's analysis procedure was extracted through structured interviews and encoded as typed functions. The model selects the tool. The tool validates the input schema, executes the analysis locally and returns deterministic results. Three researchers now use this tool for periodic structure analysis on their respective samples.

This case is included to show that the pattern generalizes beyond spectroscopy. The instrument, the file format and the analysis are different. The architectural constraint is the same: the calibration binds the tool to the physical device, and the tool must run where the device is. Commercial platforms cannot replicate this workflow: the calibration files are device-specific and the analysis requires pixel-scale parameters that only the local tool encodes. The bottom of Figure~\ref{fig:pipeline} summarizes both deployment cases.

\subsection{Application Methodology }

\noindent
{\bf Cost for interviews.}
As regards the cost of applying this approach, 
note that cost for interviews is relatively small:
From our experience, a use case requires at most three sessions of approximately thirty minutes each.

\noindent
{\bf Cost for software development.}
The cost for producing  a typed tool
comprises 
(a) the time to develop it (around 1 hour),
(b)  and time to refine and test it.
        The cost of this process does not scale with interview duration or tool assembly time. The tool itself is straightforward to implement once the interview has specified what to build. The real cost is validation: the researcher whose workflow is being automated must run the tool against real data and identify inaccuracies, which are then corrected on either the tool side or the prompt side. In the photoluminescence case (\S \ref{sec:Photo}), 
        seven of the thirty interview answers contradicted the initial implementation, each requiring a correction-and-validation cycle. The development cost therefore scales with the number of methodological corrections required, not with the duration of the interview or the assembly of the tool.
        For the Scanning Electron Microscopy case (\S \ref{sec:Scanning}), the tool comprised eight typed functions and the full cycle from interview to validated deployment took approximately one day.

\noindent
{\bf Skills file.}
As regards the skills file, writing it takes minutes. It is a configuration document, not code. The skill file is written last, after the tool has been built and validated against real data. Its purpose is to constrain the model's behavior during orchestration: which tools to call in what order, what parameter values to prefer, and how to respond when the user's prompt is underspecified or when the tool returns an ambiguous error.

The content of the skill file matters more than its length. Modern language models are trained to be helpful, which means that when an error occurs they will attempt to use their reasoning and any available tools to find alternative solutions. In a scientific workflow this is dangerous because the alternative solution may produce a plausible but incorrect result. The skill file preempts this by defining predetermined behavior for failure modes, user preferences, and workflow-specific edge cases. For example, the PL skill file (Appendix \ref{sec:Skills}) specifies exact parameter names and rejects plausible-sounding alternatives that the typed schema would not accept.

A poorly written skill file is worse than having no skill file at all. Without one, the model still has the system prompt and the tool descriptions that accompany every call, and it can operate on those alone. Recent studies of over ten thousand MCP servers confirm that the quality of tool descriptions has a measurable impact on agent tool selection and argument passing~\cite{hasan2026smelly,wang2026docs2desc}. A bad skill file actively forces the model to pick tools and processes that do not match the user's pace or requests. This is why the skill file must be tailored to the specific user and their tasks.

The methodology for producing a correct skill file is: (1) build and validate the tool against real data until the researcher confirms the outputs match their manual procedure, (2) observe where the model makes wrong choices during interactive testing with realistic prompts, and (3) write the skill to close those gaps. Testing follows the same principle: the researcher runs the tool through the model on representative datasets and checks whether the model's tool selection and parameter choices match what they would have done manually. Mismatches are corrected in the skill file and retested until the model's orchestration is stable.
\newpage
\section{Evaluating Reproducibility}
\label{sec:evaluation}

\noindent
{\bf Models Evaluated.}
To test whether typed mediation eliminates result variance in practice, we ran the same analysis on four platforms: 
(a) FORTHought, 
(b) GPT-5.5 with Extended Thinking,
(c) Claude Sonnet 4.6 with Adaptive Thinking, and 
(d) Gemini 3.1 Pro. 

\smallskip
\noindent
{\bf Evaluation Dataset and Workflow.}
We used the photoluminescence dataset from Section~\ref{sec:deploymentCases}, exported from the proprietary analysis software as a CSV file. Each platform received the same file and the same natural-language prompt asking for fits across all power levels, extraction of peak intensity and position for each power, and allometric fits of the resulting power dependence. This is the typical workflow of an experimental scientist: export data from instrument software, upload to an AI assistant, describe the desired analysis in plain language. No prompt engineering, no parameter tuning, no specialized configuration. We ran each platform four times with no modifications between runs.

\smallskip
\noindent
{\bf Evaluation Setup.}
We use four runs as a repeatability check under identical inputs, prompts and data. For the typed workflow, the analysis procedure is fixed in the tool, so repeated execution should return the same result unless the implementation or input changes. For the code-generating workflows, any change in fitting method, preprocessing step or numerical output across repeated runs shows that the procedure is being reconstructed at regeneration time. This is the risk we target: in scientific use, even one inconsistent rerun can affect which result a researcher decides to trust.

\begin{table}[H]
\caption{Reproducibility comparison across four platforms. Same dataset, same prompt, four runs each. Intensity $b$ is the exponent from the allometric fit $I = aP^b$.}
\label{tab:reproducibility}
\centering
\footnotesize
\begin{tabular}{lcccc}
\toprule
 & \textbf{FORTHought} & \textbf{GPT-5.5} & \textbf{Claude 4.6} & \textbf{Gemini 3.1} \\
\midrule
Valid results     & 4/4 & 4/4 & 4/4 & 0/4 \\
Deterministic     & 4/4 & 0/4 & 0/4 & 0/4 \\
$b$ (range)       & 1.025 & 0.899--1.000 & 1.001--1.025 & 0.237$^\dagger$ \\
$\sigma_b$        & 0.000 & 0.041 & 0.010 & --- \\
Intensity spread  & ${<}0.01$\% & 54.8\% & 30.1\% & N/A \\
Avg.\ time/run    & 0:34 & 4:21 & 5:15 & 1:50 \\
\bottomrule
\multicolumn{5}{l}{\scriptsize $^\dagger$Fit $R^2 = 0.005$; centered on incorrect spectral region.}
\end{tabular}
\end{table}

\begin{figure}[H]
  \centering
  \includegraphics[width=\columnwidth]{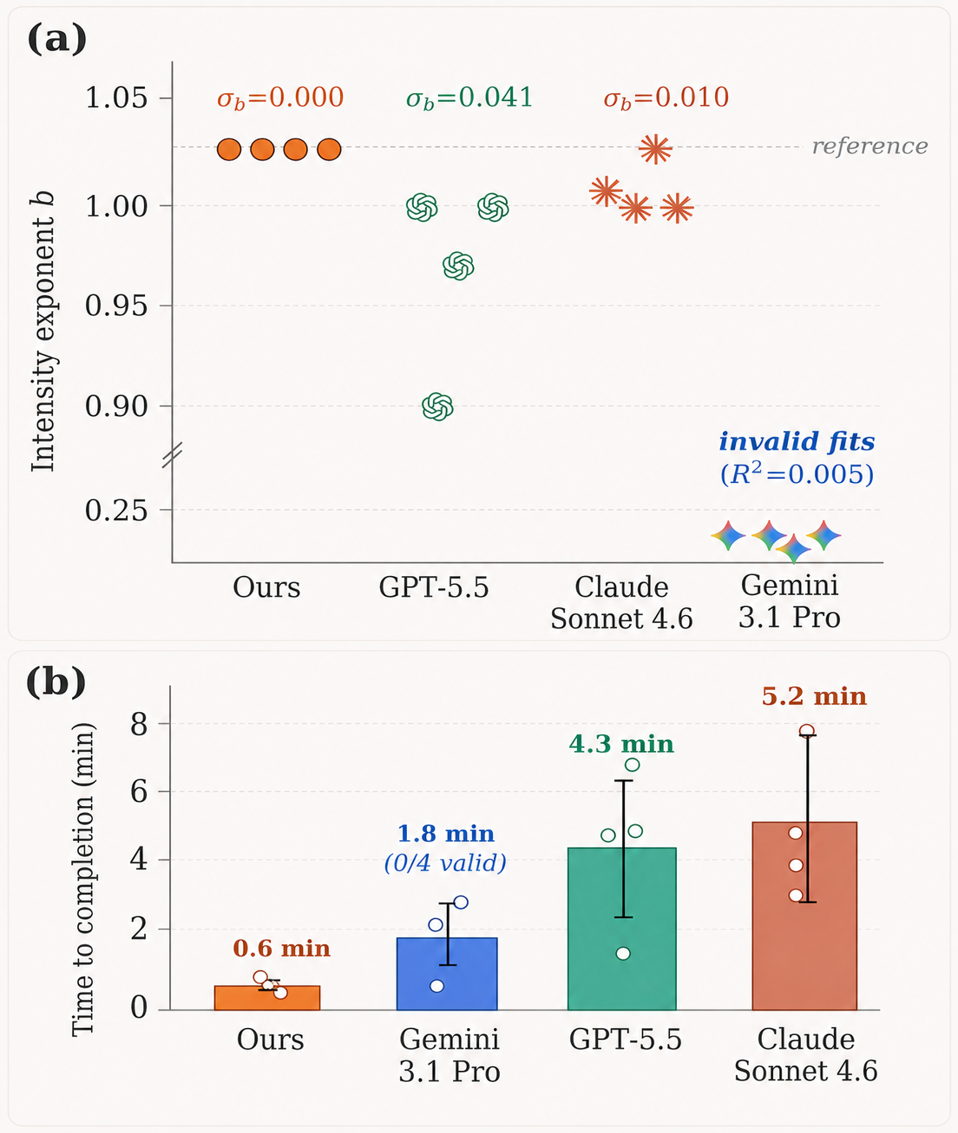}
  \caption{Reproducibility evaluation across four runs per platform.
  (a)~Intensity exponent $b$ from the allometric power-law fit. Our platform produces the same value across all runs ($\sigma_b = 0$); commercial platforms vary by an order of magnitude in spread. Gemini 3.1 returned values clustered near $b \approx 0.24$ from invalid fits ($R^2 = 0.005$); the y-axis is broken to show this. The dotted line marks the reference value from the researcher's own analytical procedure.
  (b)~Time to completion, sorted by mean. Bars show mean across four runs, error bars show one standard deviation, dots show individual runs. Gemini 3.1 produced no valid results despite finishing fastest among commercial platforms.}
  \label{fig:reproducibility}
\end{figure}

\smallskip
\noindent
\newpage
{\bf Evaluation Results.}
Table~\ref{tab:reproducibility} and Figure~\ref{fig:reproducibility} summarize the results.
We can see that:

\begin{itemize}
\item Our platform produced identical results across all four runs: $b = 1.025$ with $R^2 = 0.995$. The output files from two of the four runs were byte-for-byte identical. A third differed by one digit in the fifth decimal place of a single coefficient due to floating-point arithmetic.

\item GPT-5.5 produced valid outputs in all four runs but applied a different analytical procedure each time. Run~1 identified one emission peak. Runs 2 through 4 identified two. The background subtraction model changed between runs, with linear, quadratic and polynomial variants appearing across the four runs. The intensity of the primary peak at the lowest excitation power varied by 54.8\% across runs. These differences do not represent numerical fluctuations around a stable methodology. Each run constitutes a different experiment performed on the same data.

\item Claude Sonnet 4.6 was qualitatively more consistent, always identifying one peak, though in most runs it selected the wrong emission feature rather than the primary peak. The fit window varied from 122 to 145~nm across the four runs, with no two runs using the same bounds. Intensity values at shared power levels disagreed by up to 30.1\%.

\item Gemini 3.1 Pro failed on all four runs. Every run missed a required preprocessing step (axis unit conversion in the exported data) and produced fits with $R^2 \approx 0.005$ centered on incorrect spectral regions.
\end{itemize}

All three commercial platforms were accessed through their respective paid subscription plans with all available features enabled, including code execution environments and extended context.

Our platform also completed each run in under one minute (mean 0:34), compared to 1:50--5:15 for the commercial platforms, reflecting the absence of code generation overhead.

The typed tool does not encode the expected answer. It encodes one researcher's analytical procedure: which fitting model to use, what spectral window to fit over, what despiking parameters to apply. If the input data changes, the result changes, but the procedure stays the same. An independent Lorentzian fit (deliberately different from the tool's Voigt profile) of the same dataset using a different spectral window and no despiking produces $b = 1.036$ ($R^2 = 0.997$), which differs from the tool's output by $\Delta b = 0.011$, within the fit uncertainty of $\pm 0.027$. The value of $b$ depends on methodology. What the typed tool guarantees is that the methodology does not change between runs.

\section{Concluding Remarks}
\label{sec:CR}

In this paper
(a) we proposed a typed-mediation pattern in which both the human and the language model address the laboratory software via the same typed interface, placing the deterministic core of the workflow in the tool rather than the model,
(b) we presented two deployed applications, ported through structured interview sessions with the researchers who own the workflows, from a broader platform serving eleven active users over approximately six months, 
(c) we evaluated reproducibility by running the same analysis on our platform and three commercial foundation models and showed that the typed tool produces identical results across runs ($\sigma_b = 0$) while the commercial platforms vary in both numerical output and analytical methodology on every run, and
(d) we argued that deployment topology is a structural requirement of scientific tool mediation, 
driven by both privacy concerns and the licensing constraints of most laboratory software, which together force the tool to live alongside the data and the instrument it operates.

In the primary case, a photoluminescence analysis workflow that previously required approximately two weeks of manual processing per measurement campaign now executes in minutes and produces identical results across runs. In the second case, a scanning electron microscopy workflow for periodic structure analysis is now used by three researchers with calibration-checked results. The researcher who owns the PL workflow has used it in over sixty sessions across six months of continuous operation.

The evaluation in this work is limited to one dataset, one instrument workflow and one set of commercial platforms at a single point in time. We do not claim that code-generating approaches are inherently incapable of reproducibility, only that they did not achieve it on this task under standard usage conditions.

\noindent
{\bf Applicability.}
    In general, the proposed approach is appropriate 
    for multi-step analyses 
    where apart from the result
    we would like to have reproducibility.
The approach is most valuable whenever the task involves tooling
and the methodology, if performed incorrectly, produces silently wrong results. Modern language model deployments are almost never tool-free, even baseline interfaces provide web search, code execution and file handling. The relevant boundary is therefore not between tool-assisted and tool-free tasks, but between tasks where the methodology should be fixed and tasks where variation is acceptable. Typed mediation offers diminishing returns for tasks that are inherently exploratory (the researcher has not yet settled on a procedure to encode), tasks where generic tool use is already consistently correct, and tasks where approximate results are sufficient. It is most valuable where a non-trivial analytical procedure must be executed identically every time, which describes the majority of instrument-driven experimental workflows.

Issues for further work and research include:
(i) extending the evaluation to additional datasets and instrument types, including a user study measuring task completion time and error rates with and without typed mediation,
(ii) extending the pattern to additional instrument classes and evaluating cross-institutional deployment, and
(iii) investigating how the interview-driven encoding process itself can be partially automated as the number of ported workflows grows.

\bibliographystyle{unsrt}
\newpage
\bibliography{setn}

\appendix 

\section{An Example Skill Document for SEM Periodicity Analysis}
\label{sec:Skills}

Below is a lightly edited excerpt from the skill document for the SEM periodicity tool (Section~\ref{sec:deploymentCases}). The model reads this document before interacting with the tool. Parameter names, types, and constraints constitute the typed schema~$S$ through which the model's output is channeled.

\smallskip

\begin{tcolorbox}[colback=gray!5, colframe=gray!60, fontupper=\small, left=4pt, right=4pt, top=4pt, bottom=4pt, boxrule=0.5pt,breakable]

\textbf{SEM Analysis --- Periodicity \& Particle Sizing}

\medskip
\textbf{Workflow}
\begin{enumerate}[nosep, leftmargin=1.2em]
  \item Read magnification from the SEM info bar (e.g.\ \texttt{x40000}).
  \item Identify the uploaded file from its metadata.
  \item Decide analysis type: periodicity, particle sizing, or both.
  \item Call the tool with exact parameters.
  \item Present results with embedded figures.
\end{enumerate}

\medskip
\textbf{Analysis Type Selection}

\smallskip
{\footnotesize
\begin{tabular}{@{}p{0.50\linewidth}l@{}}
\toprule
\textbf{User request} & \textbf{\texttt{particle\_analysis}} \\
\midrule
Periodicity, spacing, LIPSS, FFT & omit (default \texttt{false}) \\
Particle/grain size, distribution & \texttt{true} \\
General ``analyze this'' & \texttt{true} (runs both) \\
\bottomrule
\end{tabular}
}

\medskip
\textbf{Tool Invocation}

\smallskip
Periodicity only:
\begin{Verbatim}[fontsize=\scriptsize, frame=lines]
run("sem_fft", {
  "file_id":  "<UUID>",
  "mag_label": "x40000"
})
\end{Verbatim}

With particle sizing:
\begin{Verbatim}[fontsize=\scriptsize, frame=lines]
run("sem_fft", {
  "file_id":  "<UUID>",
  "mag_label": "x40000",
  "particle_analysis": true
})
\end{Verbatim}

\medskip
\textbf{Schema Constraints --- Exact Parameter Names}

\smallskip
{\footnotesize
\begin{tabular}{@{}lll@{}}
\toprule
\textbf{Parameter} & \textbf{Type} & \textbf{Rejected aliases} \\
\midrule
\texttt{file\_id}            & string (UUID) & \texttt{image\_id}, \texttt{id} \\
\texttt{mag\_label}          & string        & \texttt{magnification}, \texttt{mag} \\
\texttt{particle\_analysis}  & boolean       & \texttt{analyze\_particles} \\
\bottomrule
\end{tabular}
}

\smallskip
\noindent
Extra parameters (\texttt{crop\_bottom\_px}, \texttt{roi}, \texttt{preset}) are not accepted unless the user explicitly requests them. The schema rejects any invocation that does not match these exact names and types.

\end{tcolorbox}

\smallskip
\noindent
The document instructs the model \emph{what to observe} (Step~1: read magnification), \emph{how to decide} (the analysis-type table), and \emph{how to call the tool} (exact parameter names and types). The model cannot bypass the schema: a call with \texttt{magnification} instead of \texttt{mag\_label} is rejected. This is the mechanism by which $x \in \mathcal{X}$ is enforced at runtime.

\section{Structured Interview Excerpt}
\label{sec:Interview}

The workflow encoded in each typed tool was extracted through structured interviews with the researcher who owns the procedure. Below we list a representative subset of the thirty questions used in the photoluminescence case (Section~\ref{sec:deploymentCases}). Each question targets one analytical decision. In several cases the answer contradicted the assumption we had implemented, requiring the tool to be corrected before deployment.

\smallskip

\begin{tcolorbox}[colback=gray!5, colframe=gray!60, fontupper=\small, left=4pt, right=4pt, top=4pt, bottom=4pt, boxrule=0.5pt, breakable]

{\footnotesize
\begin{enumerate}[nosep, leftmargin=1.5em, label=Q\arabic*.]

\item \textbf{What fitting profile do you use for the emission peak?}\\
\emph{Answer:} Voigt. \\
\emph{Impact:} We had implemented Lorentzian. Changed to Voigt with free Gaussian and Lorentzian widths.

\item \textbf{Is the fitting window fixed across all excitation power levels?}\\
\emph{Answer:} No. Wider at high power, narrower at low power to avoid fitting noise in the wings.\\
\emph{Impact:} We had used a single fixed window. Changed to adaptive windows.

\item \textbf{What is your R$^2$ quality threshold for accepting a fit?}\\
\emph{Answer:} Approximately 0.90, evaluated visually by overlaying the fit on the data.\\
\emph{Impact:} We had implemented a strict 0.95 cutoff. Relaxed to 0.85 to match real practice.

\item \textbf{How do you extract peak intensity?}\\
\emph{Answer:} Screen reader --- cursor placed on the peak maximum, raw counts read directly. Not from the integrated area under the fit.\\
\emph{Impact:} We had used integrated fit area. Changed the extraction metric to peak height.

\item \textbf{Do you fit a single power law across the full excitation range?}\\
\emph{Answer:} No. Two separate allometric fits ($I = aP^b$), split at a saturation boundary near ${\sim}10~\mu$W.\\
\emph{Impact:} We had implemented a single fit. Added split fitting with configurable boundary.

\item \textbf{Do you apply smoothing to the spectra before fitting?}\\
\emph{Answer:} No smoothing. Fits are performed on raw background-subtracted data.\\
\emph{Impact:} Confirmed. We had not implemented smoothing, but had considered adding it.

\item \textbf{Do you exclude outlier points from the power-law fit?}\\
\emph{Answer:} No. All surviving fit results are plotted and fitted as-is.\\
\emph{Impact:} We had implemented sigma-clipping outlier rejection. Removed it.

\item \textbf{Do you propagate fit parameters from one power level to the next as initial guesses?}\\
\emph{Answer:} No. Each spectrum is fitted independently from scratch.\\
\emph{Impact:} We retained cascade seeding as an internal optimization (improves convergence) but ensured it does not change the result.

\end{enumerate}
}

\end{tcolorbox}

\smallskip
\noindent
Each answer encodes a methodological decision that, if guessed incorrectly, changes the numerical output. The full set of thirty questions covered preprocessing (background subtraction, cosmic ray removal), model selection, quality control, parameter extraction, and post-processing. The resulting workflow specification was validated against the researcher's own manual results, achieving $\Delta \leq 0.02$ on the power-law exponent~$b$.

\end{document}